\documentclass[10pt,twocolumn,letterpaper]{article}

\usepackage{wacv}              % To produce the CAMERA-READY version
\usepackage{booktabs,tabularx,multirow}
\usepackage[section]{placeins}  
\usepackage{dblfloatfix}        
\usepackage{graphicx}
\usepackage{amsmath,amssymb,mathtools}
\usepackage[pagebackref,breaklinks,colorlinks]{hyperref}

\usepackage[capitalize]{cleveref}
\crefname{section}{Sec.}{Secs.}
\Crefname{section}{Section}{Sections}
\Crefname{table}{Table}{Tables}
\crefname{table}{Tab.}{Tabs.}

\usepackage{xcolor}

\def\wacvPaperID{*****} % *** Enter the WACV Paper ID here
\def\confName{WACV}
\def\confYear{2025}

\begin{document}

%%%%%%%%% TITLE 
\title{\textbf{DGM\textsuperscript{4}+}: Dataset Extension for Global Scene Inconsistency}

\author{Gagandeep Singh\textsuperscript{1}, Samudi Amarsinghe\textsuperscript{1}, Priyanka Singh\textsuperscript{1}, Xue Li\textsuperscript{1}
}
\maketitle
%%%%%%%%% ABSTRACT
\begin{abstract}
The rapid advances in generative models have significantly lowered the barrier to producing convincing multimodal disinformation. Fabricated images and manipulated captions increasingly co-occur to create persuasive false narratives. While the Detecting and Grounding Multi-Modal Media Manipulation (DGM\textsuperscript{4}) dataset established a foundation for research in this area, it is restricted to local manipulations such as face swaps, attribute edits, and caption changes. This leaves a critical gap: global inconsistencies, such as mismatched foregrounds and backgrounds, which are now prevalent in real-world forgeries. 

To address this, we extend DGM\textsuperscript{4} with 5,000 high-quality samples that introduce Foreground–Background (FG-BG) mismatches and their hybrids with text manipulations. Using OpenAI’s \texttt{gpt-image-1}\cite{openai2025gptimage1} and carefully designed prompts, we generate human-centric news-style images where authentic figures are placed into absurd or impossible backdrops (e.g., a teacher calmly addressing students on the surface of Mars). Captions are produced under three conditions: literal, text attribute and text split, yielding three new manipulation categories: FG-BG, FG-BG+TA, and FG-BG+TS. Quality control pipelines enforce one-to-three visible faces, perceptual hash deduplication \cite{zauner2010phash}, OCR-based text scrubbing, and realistic headline length. 

By introducing global manipulations, our extension complements existing datasets, creating a benchmark \textbf{DGM\textsuperscript{4}+} that tests detectors on both local and global reasoning. This resource is intended to strengthen evaluation of multimodal models such as HAMMER, which currently struggle with FG-BG inconsistencies. We release our \textbf{DGM\textsuperscript{4}+} dataset and generation script at \url{https://github.com/Gaganx0/DGM4plus}
\end{abstract}

%%%%%%%%% BODY TEXT
\section{Introduction}
\label{sec:intro}
Disinformation, the deliberate spread of misleading information, has shaped public perception in domains ranging from elections to public health \cite{fallis2015disinformation,fetzer2004disinformation,allcott2019trends}. Compared to misinformation, which lacks intent to deceive, disinformation exploits psychological and social biases to amplify its influence \cite{chen2023spread}. On social media, falsehoods spread faster and more widely than truths, reaching larger audiences with higher engagement \cite{vosoughi2018spread}. The accelerating realism of generative adversarial networks (GANs) \cite{goodfellow2014gan} and diffusion-based models \cite{rombach2022stable} has further amplified this risk, enabling large-scale production of fabricated media.

Detecting multimodal disinformation is especially challenging when images and captions mutually reinforce each other, creating coherent but false narratives. Benchmarks such as DGM\textsuperscript{4} \cite{shao2023dgm4} were developed to address this challenge by simulating human-centric manipulations across four types: face swaps (FS), face attributes (FA), text swaps (TS), and text attributes (TA). These manipulations are local in nature, altering a facial identity, expression, or caption sentiment. However, disinformation in practice increasingly leverages \emph{global} inconsistencies to fool consumers. Models trained solely on DGM\textsuperscript{4} struggle with such cases.

\textbf{Contributions.} In this paper, we present an extension to DGM\textsuperscript{4} that introduces foreground–background (FG-BG) manipulations. Specifically:
\begin{itemize}
    \item We generate 5,000 synthetic samples with absurd or implausible backgrounds, while preserving human-centric realism. 
    \item We provide captions in literal, sentiment-shifted (TA), and context-swapped (TS) forms, producing three new manipulation classes: FG-BG, FG-BG+TA, and FG-BG+TS.
    \item We enforce rigorous filtering: one-to-three visible faces, perceptual hashing for deduplication, and OCR scrubbing of residual text.
    \item We expand the manipulation taxonomy beyond local edits, creating a benchmark for global scene-level inconsistencies. 
\end{itemize}

This contribution equips the research community with a dataset aligned to modern disinformation tactics, strengthening evaluation of models like HAMMER and future multimodal detectors.

\section{Related Work}
\label{sec:related}

Early resources targeted unimodal manipulation: FaceForensics++ \cite{rossler2019faceforensics++} focuses on facial forgeries and has served as standard for image-only detection; LIAR \cite{wang2017liar} collects short political statements annotated for veracity, supporting text-only fact-checking research. To model cross-modal inconsistencies, NewsCLIPpings \cite{luo2021newsclippings} and COSMOS \cite{aneja2023cosmos} generate (or mine) mismatched image–text pairs to simulate out-of-context news, a pervasive real-world failure mode. Recent efforts broaden modality coverage: AV-Deepfake1M++ \cite{cai2025avdeepfake++} introduces audio-visual manipulations at scale; MFND \cite{zhu2025mfnd} incorporates sentiment/stance shifts and multitask supervision. 

DGM\textsuperscript{4} \cite{shao2023dgm4} remains the most comprehensive \emph{human-centric} benchmark coupling detection with grounding: it defines four manipulation categories—face swap (FS), face attribute (FA), text swap (TS), and text attribute (TA)—and provides token- and region-level supervision. This framing has enabled unified training targets for binary detection, type classification, and textual/visual grounding. However, all four types are fundamentally \emph{local} (identity- or attribute-level edits on the face; lexical or affective perturbations in the caption). They do not include \emph{global} scene-level inconsistencies, e.g., an otherwise authentic subject placed against an absurd or physically impossible background. Our extension complements this space by introducing foreground–background (FG–BG) mismatches—alone and combined with TA/TS—thereby probing models’ ability to reason about scene plausibility.

\subsection{Multimodal manipulation detection}
Detection methods co-evolve with benchmarks. COSMOS \cite{aneja2023cosmos} learns cross-modal alignment in a self-supervised manner via matched vs.\ mismatched pairs, effectively addressing out-of-context misuse. UFAFormer \cite{liu2025ufa} augments RGB processing with frequency-domain cues to expose manipulation artifacts, improving visual forgery sensitivity. MFND \cite{zhu2025mfnd} and ASAP \cite{zhang2024asap} exploit contrastive and semantic-alignment objectives to better couple image–text semantics. 

HAMMER \cite{shao2023dgm4} is a representative state-of-the-art detector for DGM\textsuperscript{4}: it jointly performs (i) binary manipulation detection, (ii) multi-label manipulation typing over \{FS, FA, TS, TA\}, and (iii) grounding in image regions and text tokens. This joint training yields strong performance on local manipulations seen during training. Yet, as we demonstrate, the same inductive biases and supervision that make HAMMER excel at face- and token-centric edits leave it under-specified for \emph{global} inconsistencies that require reasoning about foreground–background compatibility and scene plausibility. This motivates our FG–BG extension as a targeted stress test to complement existing evaluation regimes.

\subsection{Related directions}
Generative modeling progress (GANs \cite{goodfellow2014gan} and diffusion \cite{rombach2022stable}) has lowered the barrier to producing photorealistic manipulations, widening the space of plausible forgeries. On the societal side, misinformation dynamics \cite{fallis2015disinformation,fetzer2004disinformation,allcott2019trends,chen2023spread,vosoughi2018spread} highlight how multimodal fabrications spread and persuade at scale, highlighting the need for benchmarks that reflect contemporary deception tactics. Graph- or structure-aware perspectives on credibility \cite{monti2019fake} are complementary to our scene-plausibility focus, and could be integrated with FG–BG signals in future work.

\section{Motivation}

The motivation for this extension stems from the limitations of DGM\textsuperscript{4} and the downstream performance of models trained on it. While DGM\textsuperscript{4} covered four important manipulation types: face swaps (FS), face attributes (FA), text swaps (TS), and text attributes (TA), all of these are local in nature. They alter a person’s facial identity, appearance, or the wording of a caption, but they do not capture global inconsistencies between the subject of an image and the surrounding scene. In practice, these mismatches are increasingly common. For example, a politician addressing the press in front of an erupting volcano, or a community worker calmly standing on the surface of Mars. Such cases highlight a gap that models like HAMMER consistently fail on, since they were never exposed to this type of data during training. 

\begin{figure*}[t]
    \centering
    \includegraphics[width=\linewidth]{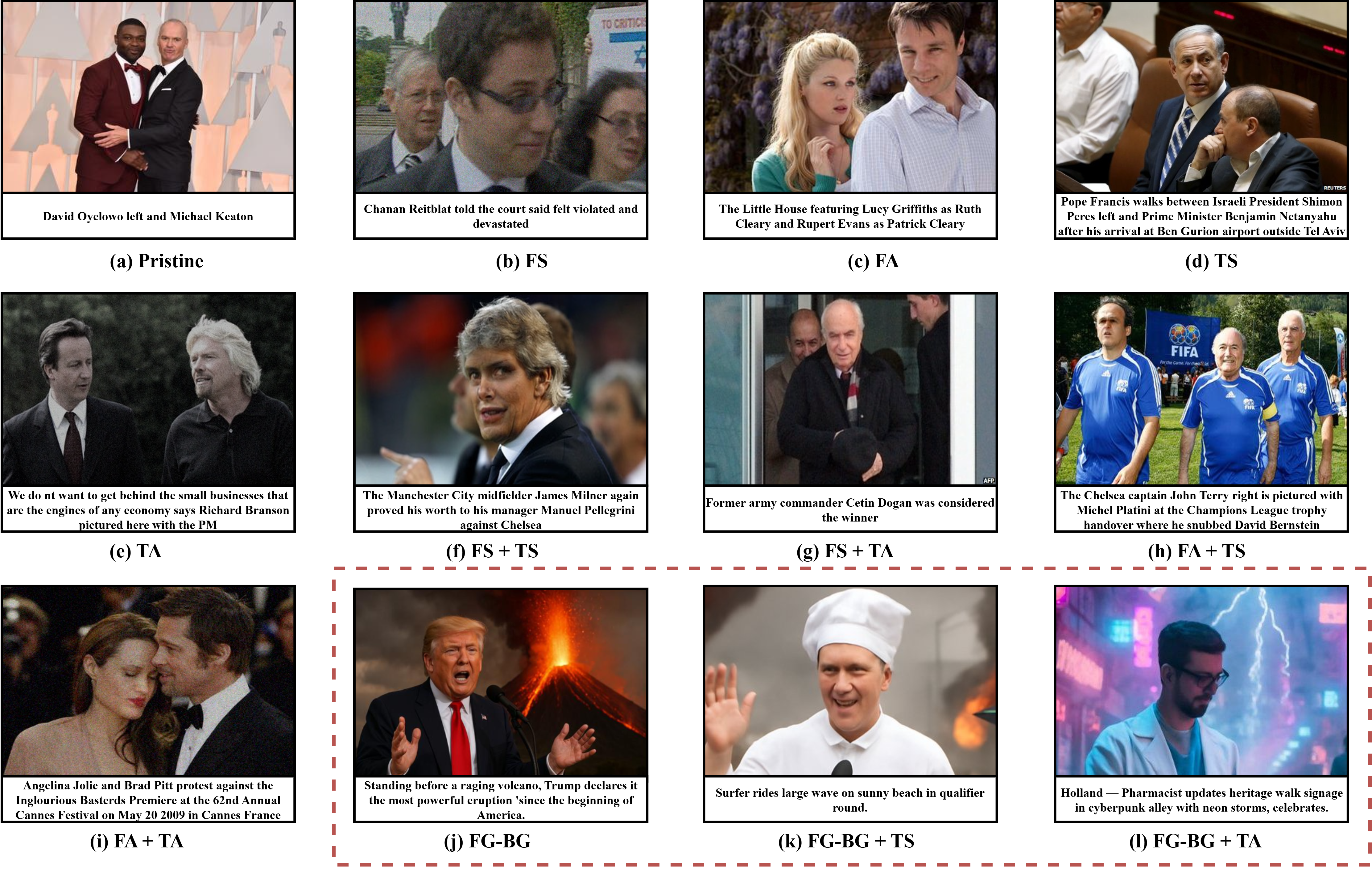} 
    \caption{Examples for each class present in the DGM\textsuperscript{4}++ dataset. The HAMMER model fails on the newly introduced three classes (j)-(l).}
    \label{fig:example-classes}
\end{figure*}

\subsection{Why current detectors (HAMMER) fail on FG--BG}
\label{sec:whyhammerfails}
HAMMER \cite{shao2023dgm4} is optimized end-to-end for the DGM\textsuperscript{4} task formulation: (1) a binary head for “manipulated vs.\ pristine,” (2) a multi-label type head defined \emph{only} over \{FS, FA, TS, TA\}, and (3) grounding heads trained with region- and token-level supervision that predominantly attends to faces and caption spans. This yields three limitations when encountering FG–BG inconsistencies:

\begin{figure*}[t]
    \centering
    \includegraphics[width=\linewidth]{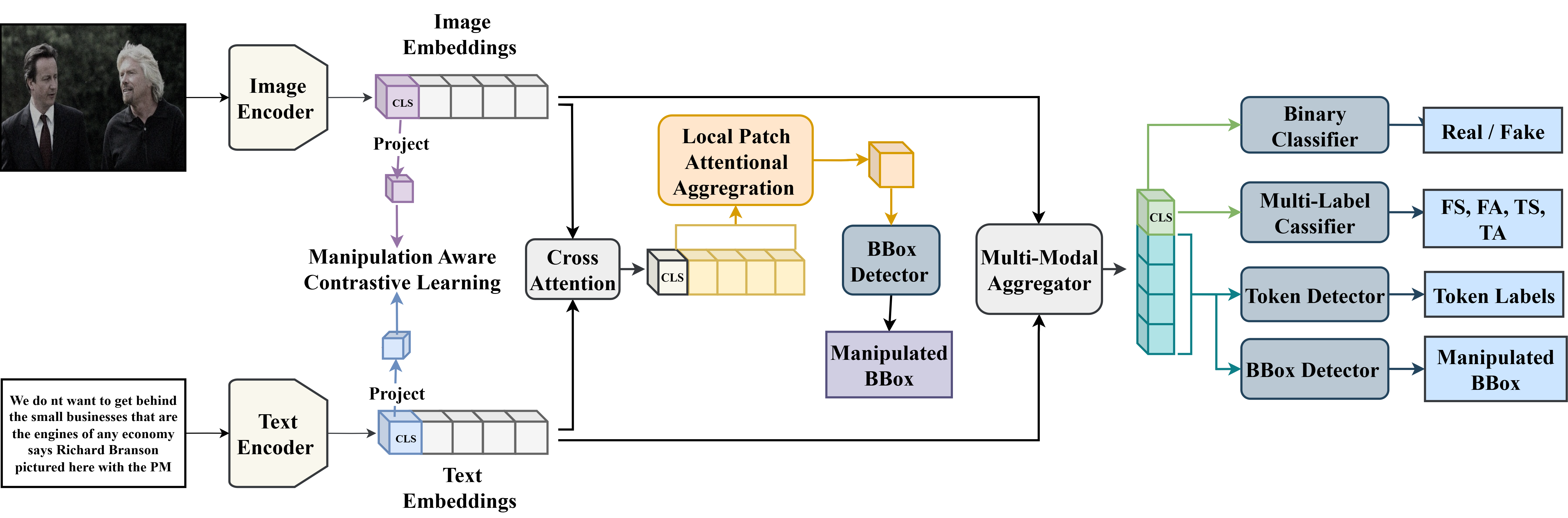} 
    \caption{Illustration of the HAMMER model. The [CLS] token of the image and text embeddings are projected into a smaller dimensionality and aligned with manipulation aware contrastive learning. The image and text embeddings are cross-attended, and the resulting patch tokens are subject to Local Patch Attentional Aggregation (LPAA) to obtain an [AGG] token, which is input to the BBox Detector for manipulation bbox detection. The image and text embeddings are input to the Multi-Modal Aggregator to obtain an aggregated multi-modal embedding. The [CLS] token of the multi-modal embedding is fed into the Binary Classifier and Multi-Label Classifier to obtain binary and fine-grained manipulation detection labels. The remainder of the aggregated tokens are input to the Token Detectors, which predict labels for each token.}
    \label{fig:hammer}
\end{figure*}

\paragraph{(A) Type-space mismatch.}
The manipulation-type classifier is explicitly defined over the DGM\textsuperscript{4} categories. FG–BG is \emph{out of vocabulary}. As a result, the multi-label head cannot express “global scene inconsistency,” and the binary head becomes the sole signal. In practice, the binary head is co-trained with type and grounding losses on local manipulations; this creates a \emph{coupling} that encourages the detector to key on local artifacts correlated with supervised types (e.g., facial identity misalignment or caption-token shifts) rather than holistic scene plausibility.

\paragraph{(B) Supervision bias toward local cues.}
Grounding supervision in DGM\textsuperscript{4} covers face boxes and text token spans \cite{shao2023dgm4}, which steers attention toward \emph{where} the known edits occur. FG–BG manipulations, by contrast, require reasoning over the \emph{compatibility} between the human subject and the entire background. Without masks, scene graphs, or foreground/background annotations, HAMMER has no training to build features that capture physical feasibility or semantic plausibility of the backdrop relative to the foreground.

\paragraph{(C) Inductive bias of objectives and data.}
Because the training set lacks FG–BG examples, the learned decision boundary favors features predictive of the four supervised types. Even though modern backbones have diverse fields, the loss never rewards discovering “absurd scene” evidence (e.g., a podium on Mars) unless it correlates with FS/FA/TS/TA. Consequently, the model can be \emph{locally correct} (e.g., no face swap detected; caption sentiment consistent) yet \emph{globally wrong} (scene is impossible). 

\paragraph{(D) Failure manifestations we target.}
Empirically, this presents as: (i) high binary confidence for “pristine” on FG–BG-only samples; (ii) diffuse or face-centric image grounding that ignores background contradictions; and (iii) over-reliance on text tokens for hybrids (FG–BG+TS/TA), where the model may latch onto TS/TA spans while still missing the scene mismatch. Our dataset extension adds labeled FG–BG cases to make global inconsistency \emph{learnable}: it decouples local token/face signals from backdrop plausibility and creates supervision to mark globally inconsistent predictions.

\paragraph{Implication.}
By adding FG–BG (and hybrids) to the training/evaluation space, we transform “scene compatibility” from an unsupervised problem to an approachable target. This enables future detectors—HAMMER variants or successors—to (1) expand the manipulation-type head to include FG–BG, (2) incorporate weak localisation (e.g., foreground/background masks) or scene descriptors, and (3) balance local forensic cues with global plausibility reasoning.

\section{Dataset Generation}
\label{sec:datasetgen}

To fill this gap, we created a controlled extension of DGM\textsuperscript{4} that introduces foreground–background (FG-BG) manipulations and their hybrids with text modifications. Our generation pipeline was designed to mimic the style and structure of newswire photographs while introducing backdrops that are semantically absurd, unsafe, or outright impossible. This ensures that the resulting samples remain visually coherent and human-centric, but are globally inconsistent in ways that test a model’s ability to reason beyond local cues.

\paragraph{Anti-shortcut design.}
We explicitly suppress easy shortcuts that arise from named entities and event references. In particular, we avoid including famous names, specific places, or well-known events in captions. This reduces the risk that a detector “latches” onto superficial cues such as city names or sports teams and instead pressures the model toward common-sense scene compatibility. We also acknowledge that news-style text is often an abstracted summary rather than a pixel-perfect description of everything in the image; our captions follow this convention to discourage brittle, token-matching heuristics.

\paragraph{Diagnostic role of mild artifacts.}
We could enforce stricter image generation and post-processing to remove visual artifacts. However, we observed that mild facial oddities, hand artifacts, and background inconsistencies usefully expose the limits of detectors that rely on local cues (for example, HAMMER’s face-centric grounding). Some samples trigger false positives by confusing identity or attributes in crowded or unusual shots, and others contain small hand anomalies that current pipelines tend to ignore. Rather than eliminating these entirely, we retain artifact-light but realistic cases to highlight and quantify such failure modes while still applying quality gates to avoid degenerate images.
\subsection{Image Generation}
\label{sec:imagegen}
We synthesize images using OpenAI’s \texttt{gpt-image-1} with \texttt{size=1024$\times$1024} and \texttt{quality="low"} (cost-efficient setting). Prompts combine three ingredients:
(1) a civic/professional \emph{role} (e.g., mayor, teacher, nurse), 
(2) a plausible \emph{news-style event} (e.g., press briefing, facility inspection),
(3) a \emph{backdrop}. 
To induce global inconsistencies, we use a curated list of $\sim$250 surreal/absurd backdrops (e.g., ``a corridor of suspended stardust'') and set \texttt{MISPLACED\_BG\_RATE=1.0}, yielding a 100\% FG--BG mismatch rate. 
Prompts explicitly discourage readable text/logo renderings and encourage unobstructed, anatomically consistent faces. 
We introduce shot diversity via \textit{wide/medium/close-up} weights.

Each image is face-aware cropped to \mbox{$400\times256$} (DGM\textsuperscript{4} target) using MTCNN\cite{zhang2016mtcnn}. We accept images with \textbf{one to three} faces and gate quality with \texttt{min\_prob=0.80} and \texttt{min\_side\_px=110}. If the gate fails, we regenerate once (\texttt{MAX\_REGEN\_ATTEMPTS=1}); otherwise the sample is discarded.

\subsection{Caption Generation}
\label{sec:capgen}
We produce three caption conditions:
\textbf{(i) Literal (Origin).} For FG–BG-only images, captions are generated by GPT-4o-mini\cite{openai2024gpt4omini} model with constrained prompting and a few illustrative examples. The instruction enforces one sentence in neutral present tense, 12–20 words, foreground subject and action first, then background introduced by “as/with/against/while,” and no locations, names, or speculative guesses. This setup mirrors real newswire tone without injecting shortcuts from named entities or events, and it intentionally abstracts rather than enumerates every visual detail.
\textbf{(ii) TA (Text Attribute).} We append an affective clause (e.g., ``appears anxious'') to the literal headline and record token spans for grounding.
\textbf{(iii) TS (Text Swap).} We replace the subject--action--place with an irrelevant yet plausible triplet drawn from curated lists.

For templated headlines we sample datelines from a 10k+ city list, conjugate subject--verb agreement, and optionally add temporal fragments (e.g., ``after audit''). This yields newswire-like sentences of 10--25 tokens.

\subsection{Filtering, Deduplication, and Safety}
\label{sec:filter}
We apply three gates:
\textbf{(a) Face gate.} MTCNN must detect $1$--$3$ faces; best face conf $\geq 0.80$; min face side $\geq 110$ px; basic eye alignment check.
\textbf{(b) OCR/text gate.} We upsample (\texttt{OCR\_SCALE=2}) and run Tesseract\cite{smith2007tesseract} with \texttt{TEXT\_CONF\_THRESH=60}. Detected words are blurred outside face boxes (Gaussian radius 6) and rechecked; residual legible text causes rejection. We blacklist high-risk tokens (e.g., institutional names) to avoid shortcuts.
\textbf{(c) Near-duplicate gate.} We compute perceptual hashes (pHash) on the final $400\times256$ image and reject any candidate within Hamming distance $\leq 3$ of existing images. We also de-duplicate captions string-exactly.

\paragraph{Incremental/resumable building.}
Before generation, we preload: counts per bucket, prior pHashes, and prior captions from existing files. We then \emph{append} JSONL records (one per sample) and write a full JSON snapshot every $N{=}50$ new samples. This prevents duplication across multiple runs and ensures crash-safe progress.

\subsection{Manipulation Taxonomy}
\label{sec:taxonomy}
Our extension adds three global-manipulation categories to DGM\textsuperscript{4}:
\textbf{FG--BG} (image-only global mismatch), 
\textbf{FG--BG+TA} (global mismatch with affective text),
\textbf{FG--BG+TS} (global mismatch with irrelevant text).
These complement the original local types (FS, FA, TS, TA) by requiring scene-level plausibility reasoning.

\section{Dataset Statistics}
\label{sec:datasetstats}

The original DGM\textsuperscript{4} dataset contained 230k image–text pairs, with 77,426 pristine samples and 152,574 manipulated samples. Our extension adds 5,000 new examples, bringing the total to 235k. Although this represents only a modest increase in scale (around 2.1 percent), it introduces a type of manipulation that was previously missing and which we argue is critical for evaluating multimodal forgery detectors.

The extension is balanced across the three new categories, with 2,000 FG-BG samples, 1,500 FG-BG+TA samples, and 1,500 FG-BG+TS samples. Captions range between 10 and 25 tokens, mirroring the length of authentic news headlines. All images respect the one-to-three face constraint and have been cropped to the standard resolution.  

\begin{figure}[t]
    \centering
    \includegraphics[width=\linewidth]{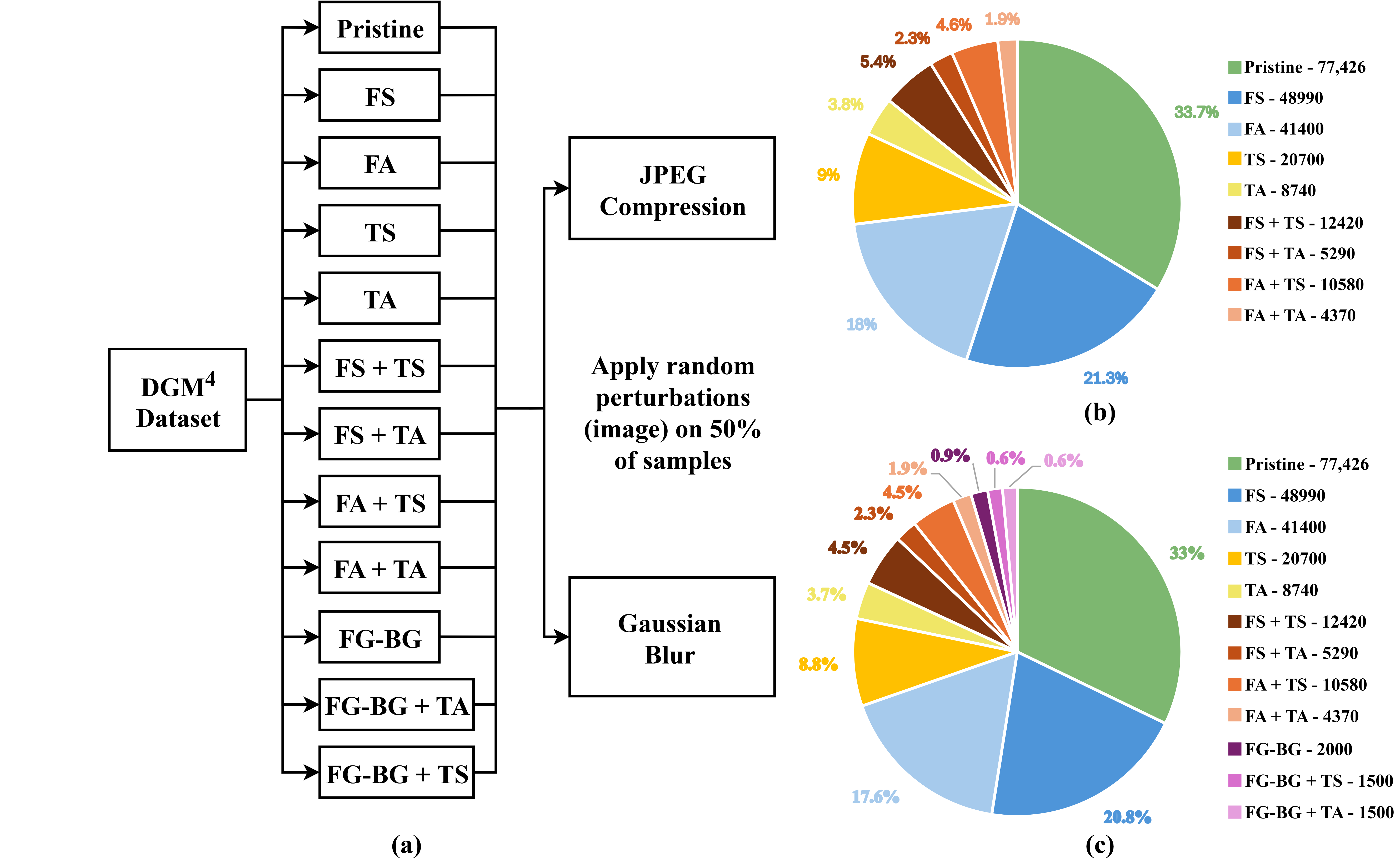}
    \caption{Distribution of the extended dataset across pristine and manipulation categories. The additional 5,000 samples cover the Foreground-Background class and its hybrids. (a) Dataset class types; (b) Original DGM\textsuperscript{4} dataset distribution; (c) Extended DGM\textsuperscript{4}++ distribution}
    \label{fig:dataset-scale}
\end{figure}

Figure~\ref{fig:dataset-scale} shows the updated class distribution. Compared to the heavy concentration of local manipulations in DGM\textsuperscript{4}, our contribution specifically targets the global FG-BG axis. These new cases include, for example, a community spokesperson calmly addressing reporters in the middle of a wildfire tornado, or a school principal described as “appearing anxious” while standing on the rings of Saturn. Such additions are designed to challenge detectors in ways that local edits cannot, forcing them to consider the plausibility of an entire scene rather than isolated features.

\section{Experiments}
\label{sec:experiments}

\paragraph{Why these baselines?}
We select four complementary families to probe FG--BG mismatch:  
(i) \textbf{Contrastive VLM encoders} (\textit{OpenCLIP}\cite{cherti2023openclip}, \textit{SigLIP})\cite{zhai2023siglip} test whether role text aligns more with the foreground than the background, operationalizing global scene compatibility as a simple similarity \emph{gap} without full reasoning;  
(ii) \textbf{Vision-only self-supervised features} (\textit{DINOv2})\cite{oquab2023dinov2} assess whether foreground and background lie far apart in feature space, isolating a purely visual signal that ignores captions;  
(iii) \textbf{Multimodal LLM} (\textit{Qwen2-VL-2B-Instruct})\cite{qwen2vl2024} answers a strict Yes/No plausibility question, offering a reasoning-style signal;  
(iv) \textbf{Task-specific detector} (\textit{HAMMER}) anchors results w.r.t.\ the widely used DGM\textsuperscript{4} model family trained on local manipulations (FS/FA/TS/TA).

\paragraph{Evaluation setup.}
We evaluate on the Extended DGM\textsuperscript{4} \emph{inconsistent} split (all images intentionally FG--BG mismatched). For one-class settings we report \emph{Acc} and \emph{F1} against all-1 labels plus the fraction flagged as inconsistent.  
Foreground masks are obtained with YOLOv8-seg\cite{ultralytics2023yolov8} by unioning up to three \texttt{person} instances; background is the complement. Contrastive encoders compute
\(
\Delta = s(\mathrm{FG}, \texttt{text}) - s(\mathrm{BG}, \texttt{text})
\),
predicting ``inconsistent'' if $\Delta < 0$.  
Qwen2-VL is queried with a binary prompt, mapping \texttt{No} to ``inconsistent''.  
HAMMER is run with its public checkpoint; FG--BG is out-of-vocabulary for its type head.

\paragraph{DINOv2 baseline (vision-only).}
To isolate visual mismatch, we add a caption-free baseline using DINOv2 (\texttt{vit\_base\_patch14\_dinov2}). Foreground and background crops are encoded, and the L2 distance between normalized features is computed. Larger distance implies greater inconsistency. This requires only FG/BG crops and outputs \texttt{id, fgbg\_dist}. With labels, we evaluate a simple proxy: threshold at the dataset median.

% ===== Table A: Vision–Language Encoders (Δ-sim, one-class)
\begin{table}[!tb]
\centering
\caption{Vision--Language encoders on the all-inconsistent split (one-class). $\Delta = s(\text{FG},\text{text}) - s(\text{BG},\text{text})$.}
\label{tab:vl}
\vspace{-2mm}
\setlength{\tabcolsep}{4pt}\scriptsize
\begin{tabularx}{\linewidth}{@{}lccc@{}}
\toprule
\textbf{Method} & $\Delta$ mean$\pm$sd & \% $\Delta<0$ & Acc / F1 \\
\midrule
OpenCLIP (ViT-B/32) & $-0.046\pm0.070$ & 74.3 & 0.743 / 0.852 \\
SigLIP (base/16)    & $-0.003\pm0.026$ & 53.9 & 0.539 / 0.701 \\
\bottomrule
\end{tabularx}
\end{table}

% ===== Table B: Vision-only distance (DINOv2)
\begin{table}[!tb]
\centering
\caption{Vision-only baseline (DINOv2 ViT-B/14). FG–BG feature distance $d=\lVert F-B\rVert_2$. One-class proxy threshold at median.}
\label{tab:dino}
\vspace{-2mm}
\setlength{\tabcolsep}{4pt}\scriptsize
\begin{tabularx}{\linewidth}{@{}lcc@{}}
\toprule
\textbf{Method} & Dist (mean$\pm$sd) & Acc / F1 @ median \\
\midrule
DINOv2 (ViT-B/14) & $1.244\pm0.030$ & 0.500 / 0.667 \\
\bottomrule
\end{tabularx}
\end{table}

% ===== Table C: Multimodal LLM (Yes/No)
\begin{table}[!tb]
\centering
\caption{Qwen2-VL-2B-Instruct with Yes/No prompt (V2 framing).}
\label{tab:vlm}
\vspace{-2mm}
\setlength{\tabcolsep}{4pt}\scriptsize
\begin{tabularx}{\linewidth}{@{}lccc@{}}
\toprule
\textbf{Method} & \% ``No'' & Acc & F1 \\
\midrule
Qwen2-VL-2B-Instruct & 32.8 & 0.328 & 0.494 \\
\bottomrule
\end{tabularx}
\end{table}

% ===== Table D: Task-specific anchor (HAMMER)
\begin{table}[!tb]
\centering
\caption{HAMMER on FG--BG extension. FG--BG is out-of-vocabulary for its type head.}
\label{tab:hammer}
\vspace{-2mm}
\setlength{\tabcolsep}{3pt}\scriptsize
\begin{tabularx}{\linewidth}{@{}lcccc@{}}
\toprule
\textbf{Method} & ACC$_{cls}$ & OF1 & IoU mean & Tok Acc / F1 \\
\midrule
HAMMER (released) & 19.1 & 35.3 & 94.9 & 78.8 / 22.0 \\
\bottomrule
\end{tabularx}
\end{table}

\section{Discussion}
\label{sec:discussion}

The experiments highlight several distinct patterns.

\paragraph{Contrastive encoders pick up global mismatch without training.}
OpenCLIP and SigLIP, trained only for generic vision–language alignment, already expose a signal: captions describing human roles align more closely with FG than BG. The sign of the similarity gap alone distinguishes many inconsistent cases. However, their performance diverges: OpenCLIP achieves strong separation, while SigLIP is much weaker. This suggests that embedding geometry, not just scale or training data, strongly influences sensitivity to global scene plausibility.

\paragraph{Vision-only features reveal stable FG--BG separation.}
DINOv2 shows that foreground and background features occupy measurably different regions in representation space, even without text. The distances are tightly distributed, hinting at a consistent internal notion of ``subject vs.\ context.'' Although our naive threshold gives only balanced accuracy, the high F1 indicates that with a calibrated cutoff the model could reliably surface inconsistencies. This supports the idea that visual-only FG–BG statistics are a valuable complement to text-aware methods.

\paragraph{VLMs struggle with binary plausibility.}
Qwen2-VL demonstrates how brittle large multimodal models can be: their outputs depend heavily on prompt framing. Despite having access to the full image and text, the model under-reports inconsistency. This optimism bias reflects how such models are tuned for helpfulness and plausibility, not adversarial detection. Structured prompts or chain-of-thought explanations may be necessary to coax more reliable judgments.

\paragraph{HAMMER reveals a fundamental supervision gap.}
HAMMER excels at tasks it was trained for—faces and token-level manipulations—but flounders on FG–BG. Its high IoU scores show it continues to ground people well, yet its binary/type heads give low accuracy because FG–BG does not exist in its type space. This demonstrates the importance of supervision: detectors trained only on local manipulations are not equipped to generalize to global ones.

\paragraph{Overall implications.}
Taken together, the results show that global inconsistency is not intractable: even off-the-shelf encoders and features carry signal. But bridging the gap requires supervision that makes FG–BG ``visible'' to detectors. Future models should integrate three pillars: (1) vision-only FG–BG distance cues, (2) contrastive alignment gaps, and (3) structured VLM reasoning. Training regimes that combine these cues with explicit FG–BG supervision could enable detectors to handle both local edits and global plausibility in a unified way.

\section{Limitations and Future Directions}
\begin{itemize}
    \item
We used median thresholding for DINOv2 to avoid peeking; a small calibration split could yield more realistic operating points and AUROC/AUPRC. For VLMs, structured few-shot rationales and tool-use (explicit FG/BG inspection) should reduce optimism bias. For HAMMER, extending the type head with FG–BG and adding weak FG/BG masks (from our YOLOv8-seg union) will align supervision with the target phenomenon and likely close most of the gap. Finally, a joint training objective that includes a \emph{contrastive FG-vs-BG alignment term} alongside standard manipulation losses is a promising direction to unify local and global cues.

    \item Limitations:
Built synthetically from a fixed prompt vocabulary, the extension may carry style and coverage biases. Despite automated QC (MTCNN face counts, OCR scrub, pHash de-dup), edge cases can slip through—or valid samples be over-filtered. Because FG–BG is annotated only at the image level, precise localization is out of scope. The 1–3-face constraint and 400×256 crop further bias the data toward human-centric, medium/close shots. Finally, our experiments to date emphasise HAMMER; broader SOTA coverage remains a future work.

    \item Future Directions:
We will broaden coverage across roles, locales, and languages and stage the benchmark with graded difficulty. We plan to add human plausibility ratings and provide weak localization for FG–BG (e.g., foreground/background masks or scene tags). Beyond single images, we will extend to video and multi-image narratives and incorporate external knowledge cues. On the modeling side, we will release a unified baseline suite—spanning joint-embedding and fusion transformers—under a common evaluation harness. Finally, we will expand fairness/QA checks and publish richer generation metadata.
\end{itemize}

\section{Ethical Considerations}
\begin{itemize}
    \item \textbf{Data collection:} All content is synthetically generated; no real images are ingested. The generator enforces OCR at creation time and blurs detected legible text, rejecting images if text remains readable. Stored annotations (boxes, token indices) refer only to synthetic content; no real-person identifiers are collected.
    \item \textbf{IP rights:} Prompts explicitly request brandless scenes, no logos/words, reducing exposure to trademarked signage and textual marks.
    \item \textbf{Misuse and disinformation risk:} Although intended to detect multimodal inconsistencies, any media dataset can be misused to refine generative systems. To reduce this risk, we (i) release clear manipulation labels and metadata for detection research, (ii) discourage repurposing for content creation, and (iii) recommend terms of use forbidding generation or dissemination of deceptive media and requiring responsible disclosure of evaluation results.
\end{itemize}
\section{Conclusions}
\label{sec:conclusions}

In this work, we have introduced an extension to the DGM\textsuperscript{4} dataset, denoted as \textbf{DGM\textsuperscript{4}+}, which adds global scene-level inconsistencies to complement existing local manipulations. This new dataset is publicly available at . By releasing this resource, we aim to provide the community with a challenging benchmark that more faithfully reflects modern multimodal disinformation scenarios.

The significance of \textbf{DGM\textsuperscript{4}+} lies in its ability to probe reasoning beyond face- and token-level manipulations, specifically addressing the limitations of current detectors such as HAMMER. Our results demonstrate that although existing baselines capture partial signals, explicit supervision over foreground–background mismatches remains essential. We expect this dataset to serve as a catalyst for future research on global plausibility, ultimately strengthening model robustness against real-world forgeries.

\section*{Acknowledgement}

We would like to acknowledge the original authors of DGM\textsuperscript{4} \cite{shao2023dgm4} for releasing their dataset, which has provided the foundation for this extension.

{\small
\bibliographystyle{ieee_fullname}
\bibliography{egbib}

\begin{thebibliography}{10}\itemsep=-1pt

\bibitem{allcott2019trends}
Hunt Allcott, Matthew Gentzkow, and Chuan Yu.
\newblock Trends in the diffusion of misinformation on social media.
\newblock {\em Research \& Politics}, 6(2):1--8, 2019.

\bibitem{aneja2023cosmos}
Shivangi Aneja, Chris Bregler, and Matthias Nie{\ss}ner.
\newblock {COSMOS}: Catching out-of-context image misuse with self-supervised learning.
\newblock In {\em AAAI Conf. Artif. Intell.}, pages 13584--13592, 2023.

\bibitem{cai2025avdeepfake++}
Zhixi Cai, Kartik Kuckreja, Shreya Ghosh, Akanksha Chuchra, Muhammad~Haris Khan, Usman Tariq, Tom Gedeon, and Abhinav Dhall.
\newblock Av-deepfake1m++: A large-scale audio-visual deepfake benchmark with real-world perturbations, 2025.

\bibitem{chen2023spread}
Sijing Chen, Lu Xiao, and Akit Kumar.
\newblock Spread of misinformation on social media: What contributes to it and how to combat it.
\newblock {\em Computers in Human Behavior}, 141:107643, 2023.

\bibitem{cherti2023openclip}
Mehdi Cherti, Romain Beaumont, Ross Wightman, Mitchell Wortsman, Gabriel Ilharco, Cade Gordon, Christoph Schuhmann, Ludwig Schmidt, and Jenia Jitsev.
\newblock Reproducible scaling laws for contrastive language-image learning.
\newblock In {\em IEEE Conf. Comput. Vis. Pattern Recognit.}, pages 2818--2829, 2023.

\bibitem{fallis2015disinformation}
Don Fallis.
\newblock What is disinformation?
\newblock {\em Library Trends}, 63(3):401--426, 2015.

\bibitem{fetzer2004disinformation}
James~H. Fetzer.
\newblock Disinformation: The use of false information.
\newblock {\em Minds and Machines}, 14(2):231--240, 2004.

\bibitem{goodfellow2014gan}
Ian Goodfellow, Jean Pouget-Abadie, Mehdi Mirza, Bing Xu, David Warde-Farley, Sherjil Ozair, Aaron Courville, and Yoshua Bengio.
\newblock Generative adversarial nets.
\newblock In {\em Adv. Neural Inf. Process. Syst.}, pages 2672--2680, 2014.

\bibitem{liu2025ufa}
Huan Liu, Zichang Tan, Qiang Chen, Yunchao Wei, Yao Zhao, and Jingdong Wang.
\newblock Unified frequency-assisted transformer framework for detecting and grounding multi-modal manipulation.
\newblock {\em Int. J. Comput. Vis.}, 133(3):1392--1409, 2025.

\bibitem{luo2021newsclippings}
Grace Luo, Trevor Darrell, and Anna Rohrbach.
\newblock {N}ews{CLIP}pings: Automatic generation of out-of-context multimodal media.
\newblock In {\em Proceedings of the 2021 Conference on Empirical Methods in Natural Language Processing}, pages 6801--6817, Online and Punta Cana, Dominican Republic, Nov. 2021. Association for Computational Linguistics.

\bibitem{monti2019fake}
Federico Monti, Fabrizio Frasca, Davide Eynard, Damon Mannion, and Michael~M. Bronstein.
\newblock Fake news detection on social media using geometric deep learning.
\newblock {\em arXiv preprint arXiv:1902.06673}, 2019.

\bibitem{openai2024gpt4omini}
{OpenAI}.
\newblock Gpt-4o mini: Advancing cost-efficient intelligence.
\newblock \url{https://openai.com/index/gpt-4o-mini-advancing-cost-efficient-intelligence/}, 2024.
\newblock Accessed 2025-09-30.

\bibitem{openai2025gptimage1}
{OpenAI}.
\newblock Our latest image generation model in the api via \texttt{gpt-image-1}.
\newblock \url{https://openai.com/index/image-generation-api/}, 2025.
\newblock Accessed 2025-09-30.

\bibitem{oquab2023dinov2}
Maxime Oquab, Timoth{\'e}e Darcet, Th{\'e}o Moutakanni, Huy~V. Vo, Marc Szafraniec, Vasil Khalidov, Pierre Fernandez, Daniel HAZIZA, Francisco Massa, Alaaeldin El-Nouby, Mido Assran, Nicolas Ballas, Wojciech Galuba, Russell Howes, Po-Yao Huang, Shang-Wen Li, Ishan Misra, Michael Rabbat, Vasu Sharma, Gabriel Synnaeve, Hu Xu, Herve Jegou, Julien Mairal, Patrick Labatut, Armand Joulin, and Piotr Bojanowski.
\newblock {DINO}v2: Learning robust visual features without supervision.
\newblock {\em Transactions on Machine Learning Research}, 2024.
\newblock Featured Certification.

\bibitem{rombach2022stable}
Robin Rombach, Andreas Blattmann, Dominik Lorenz, Patrick Esser, and Bj{\"o}rn Ommer.
\newblock High-resolution image synthesis with latent diffusion models.
\newblock In {\em IEEE Conf. Comput. Vis. Pattern Recognit.}, pages 10684--10695, 2022.

\bibitem{rossler2019faceforensics++}
Andreas R{\"o}{\ss}ler, Davide Cozzolino, Luisa Verdoliva, Christian Riess, Justus Thies, and Matthias Nie{\ss}ner.
\newblock Faceforensics++: Learning to detect manipulated facial images.
\newblock In {\em Int. Conf. Comput. Vis.}, pages 1--11, 2019.

\bibitem{shao2023dgm4}
Rui Shao, Tianxing Wu, and Ziwei Liu.
\newblock Detecting and grounding multi-modal media manipulation.
\newblock In {\em IEEE Conf. Comput. Vis. Pattern Recognit.}, pages 6904--6913, 2023.

\bibitem{smith2007tesseract}
Ray Smith.
\newblock An overview of the tesseract ocr engine.
\newblock In {\em Proc. Int. Conf. Document Analysis and Recognition (ICDAR)}, pages 629--633, 2007.

\bibitem{ultralytics2023yolov8}
Ultralytics.
\newblock {YOLOv8}: Ultralytics real-time object detection, 2023.
\newblock Code repository.

\bibitem{vosoughi2018spread}
Soroush Vosoughi, Deb Roy, and Sinan Aral.
\newblock The spread of true and false news online.
\newblock {\em Science}, 359(6380):1146--1151, 2018.

\bibitem{qwen2vl2024}
Peng Wang, Shuai Bai, Sinan Tan, Shijie Wang, Zhihao Fan, Jinze Bai, Keqin Chen, Xuejing Liu, Jialin Wang, Wenbin Ge, Yang Fan, Kai Dang, Mengfei Du, Xuancheng Ren, Rui Men, Dayiheng Liu, Chang Zhou, Jingren Zhou, and Junyang Lin.
\newblock Qwen2-vl: Enhancing vision-language model's perception of the world at any resolution, 2024.

\bibitem{wang2017liar}
William~Y. Wang.
\newblock ``liar, liar pants on fire'': A new benchmark dataset for fake news detection.
\newblock In {\em Proceedings of ACL (Short Papers)}, pages 422--426, 2017.

\bibitem{zauner2010phash}
Christoph Zauner.
\newblock Implementation and benchmarking of perceptual image hash functions.
\newblock Master's thesis, University of Applied Sciences Upper Austria, 2010.

\bibitem{zhai2023siglip}
Xiaohua Zhai, Basil Mustafa, Alexander Kolesnikov, and Lucas Beyer.
\newblock Sigmoid loss for language image pre-training.
\newblock {\em arXiv preprint arXiv:2303.15343}, 2023.

\bibitem{zhang2016mtcnn}
Kaipeng Zhang, Zhanpeng Zhang, Zhifeng Li, and Yu Qiao.
\newblock Joint face detection and alignment using multi-task cascaded convolutional networks.
\newblock {\em arXiv preprint arXiv:1604.02878}, 2016.

\bibitem{zhang2024asap}
Zhenxing Zhang, Yaxiong Wang, Lechao Cheng, Zhun Zhong, Dan Guo, and Meng Wang.
\newblock Asap: Advancing semantic alignment promotes multi-modal manipulation detecting and grounding, 2024.

\bibitem{zhu2025mfnd}
Ye Zhu, Yunan Wang, and Zitong Yu.
\newblock Multimodal fake news detection: Mfnd dataset and shallow-deep multitask learning, 2025.

\end{thebibliography}
}

\end{document}